\documentclass{article}

\PassOptionsToPackage{numbers, compress}{natbib}
\usepackage[preprint]{neurips_2025}

\usepackage[whole]{bxcjkjatype}
\usepackage[utf8]{inputenc} 
\usepackage[T1]{fontenc}    
\usepackage{hyperref}       
\usepackage{url}            
\usepackage{booktabs}       
\usepackage{amsfonts}       
\usepackage{nicefrac}       
\usepackage{microtype}      
\usepackage{xcolor}         
\usepackage{amsmath}

\PassOptionsToPackage{numbers, compress}{natbib}
\usepackage{colortbl}
\usepackage{graphicx}
\usepackage{wrapfig}
\usepackage{hyperref}
\hypersetup{colorlinks,linkcolor={red},citecolor={blue}, urlcolor=orange}  
\definecolor{COLOR_MEAN}{HTML}{f0f0f0}
\definecolor{LIGHT_BLUE}{HTML}{e6f1fe}
\definecolor{LIGHT_RED}{HTML}{fceeee}
\definecolor{LIGHT_YELLOW}{HTML}{f1f58a}
\definecolor{LIGHT_GREEN}{HTML}{eaffea}
\definecolor{LIGHT_BROWN}{HTML}{f5e6d3}
\usepackage{multirow} 
\usepackage{makecell} 
\usepackage{caption}
\usepackage{cleveref}
\usepackage{tcolorbox}
\usepackage[subrefformat=parens]{subcaption}
\tcbuselibrary{breakable}
\usepackage{fontawesome}
\usepackage{url}

\definecolor{darkblue}{RGB}{84, 112, 198}
\definecolor{lightgreen}{RGB}{145, 204, 117}
\definecolor{lightyellow}{RGB}{250, 200, 88}
\definecolor{lightred}{RGB}{238, 102, 102}
\definecolor{lightblue}{RGB}{115, 192, 222}

\newtcolorbox{promptbox}[2][Prompt]{
colback=black!5!white,
arc=5pt, 
boxrule=0.5pt,
fonttitle=\bfseries,
title=#1, 
before upper={\small}, fontupper=\fontfamily{ptm}\selectfont,
colframe=#2,
}

\newcommand{\eg}{\textit{e.g.,}}
\newcommand{\ie}{\textit{i.e.,}}

\title{\fontsize{14pt}{15pt}\selectfont JMMMU-Pro: Image-based Japanese Multi-discipline Multimodal Understanding Benchmark via Vibe Benchmark Construction}

\author{%
Atsuyuki Miyai \quad
Shota Onohara \quad
Jeonghun Baek \quad
Kiyoharu Aizawa
\\[1mm]
\texttt{miyai@cvm.t.u-tokyo.ac.jp}\\
{\texttt{\{onohara, baek, aizawa\}@hal.t.u-tokyo.ac.jp}}
\\[1mm]
The University of Tokyo
\\[1.5mm]
\faHome\ \url{https://mmmu-japanese-benchmark.github.io/JMMMU_Pro/}\\
\vspace{-1pt}
\faDatabase\ \href{https://huggingface.co/datasets/JMMMU/JMMMU-Pro}{Dataset}\quad
\faGithub\ \href{https://github.com/EvolvingLMMs-Lab/lmms-eval}{Code}\quad
\faTrophy\ \href{https://huggingface.co/spaces/JMMMU/JMMMU-Pro_Leaderboard}{Leaderboard}
}

\begin{document}

\maketitle
\begin{figure*}[h]
\vspace{-25pt}
\centering
    \includegraphics[width=0.99\linewidth]{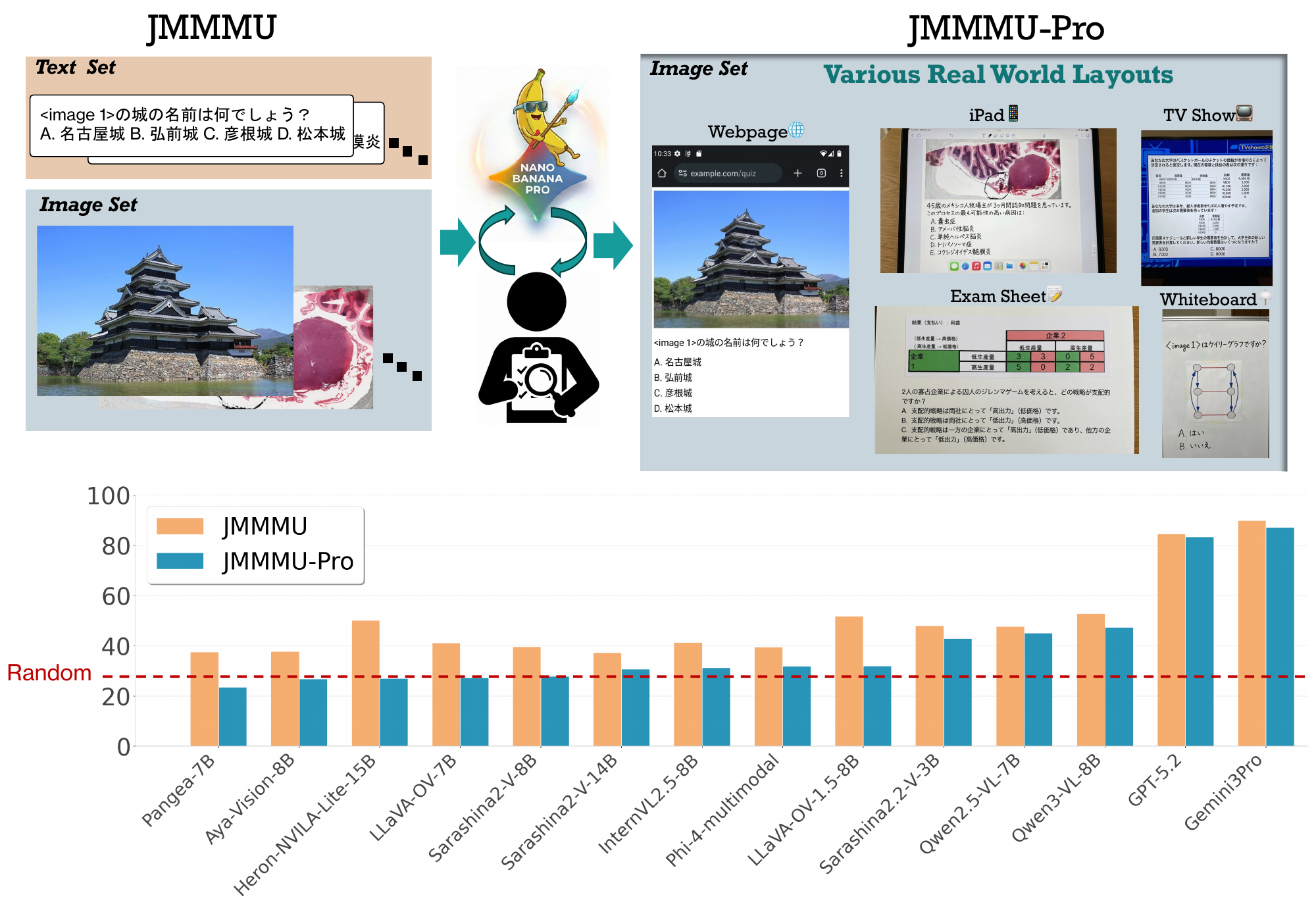}\\
    \vspace{-5pt}
   \caption{\textbf{Building JMMMU-Pro via Vibe Benchmark Construction}. JMMMU-Pro extends JMMMU by embedding each question image and text into a single image. To construct JMMMU-Pro, we propose Vibe Benchmark Construction, where an image generation model creates questions, followed by human verification and prompt refinement to ensure quality. Experiments indicate that current open-source LMMs struggle with JMMMU-Pro.}
    \label{fig:fig_teaser}
\end{figure*}

\begin{abstract}
    This paper introduces JMMMU-Pro, an image-based Japanese Multi-discipline Multimodal Understanding Benchmark, and Vibe Benchmark Construction, a scalable construction method. Following the evolution from MMMU to MMMU-Pro, JMMMU-Pro extends JMMMU by composing the question image and question text into a single image, thereby creating a benchmark that requires integrated visual-textual understanding through visual perception. To build JMMMU-Pro, we propose Vibe Benchmark Construction, a methodology in which an image generative model (\eg~Nano Banana Pro) produces candidate visual questions, and humans verify the outputs and, when necessary, regenerate with adjusted prompts to ensure quality. By leveraging Nano Banana Pro's highly realistic image generation capabilities and its ability to embed clean Japanese text, we construct a high-quality benchmark at low cost, covering a wide range of background and layout designs. Experimental results show that all open-source LMMs struggle substantially with JMMMU-Pro, underscoring JMMMU-Pro as an important benchmark for guiding future efforts in the open-source community. We believe that JMMMU-Pro provides a more rigorous evaluation tool for assessing the Japanese capabilities of LMMs and that our Vibe Benchmark Construction also offers an efficient guideline for future development of image-based VQA benchmarks.
\end{abstract}
\section{Introduction}
With the recent success of large multimodal models (LMMs) in English~\citep{gpt4o, liu2024improved, liu2024llavanext}, there has been a growing interest in developing multilingual LMMs~\citep{wang2024qwen2, dash2025aya, yue2024pangea, Bai2025Qwen3VLTR} and LMMs specialized for non-English languages~\citep{sbintuitions_sarashina2.2_vision_3b, team2025hyperclova}. Although LMM development in the Japanese domain has emerged~\citep{sbintuitions_sarashina2.2_vision_3b, baek2025harnessing, sasagawa-etal-2025-constructing}, progress has been slower than in the English domain, in part due to the limited evaluation benchmarks.
Given the large and rapidly growing population of Japanese LMM users, there is an increasing need to establish more Japanese benchmarks that can facilitate the development of LMMs capable of handling the Japanese language and culture seamlessly.

Among the several benchmarks for Japanese LMMs~\citep{inoue2024heron, baek2025mangavqa, onami2024jdocqa, onohara2025jmmmu}, one of the most representative is JMMMU (Japanese Massive Multi-discipline Multimodal Understanding Benchmark)~\citep{onohara2025jmmmu}. Inspired by the MMMU benchmark~\citep{yue2024mmmu}, JMMMU is the first benchmark designed to evaluate LMMs on extensive, multi-disciplinary tasks in Japanese that require college-level subject knowledge, deliberate reasoning, and cultural understanding.
JMMMU consists of a culture-agnostic (CA) subset of 720 items, constructed through translation from MMMU, and a culture-specific (CS) subset of 600 items that incorporate Japanese cultural elements. This systematic design enables apple-to-apple comparisons with the original MMMU through the CA subset while simultaneously evaluating cultural understanding through the CS subset. Due to its comprehensive and rigorous evaluation coverage, JMMMU has become a foundational benchmark for the development of Japanese LMMs~\citep{sbintuitions_sarashina2.2_vision_3b, sbintuitions_sarashina2_vision_14b, turingmotors_heron_nvila_lite_15b}.

A major limitation of existing Japanese benchmarks is that the question image and the question text are provided to the model as separate modalities. This evaluation setup differs substantially from the core human cognitive skill: \textit{Seamlessly integrating visual and textual information and interpreting them through visual perception}. Equipping LMMs with this cognitive ability in Japanese is a crucial step toward developing embodied agents and robotic systems~\cite{zitkovich2023rt2, ichter2022do, huang2022zeroshot, li2024embodiedinterface} that can autonomously operate and explore real-world environments in Japan through visual perception.
Furthermore, from the perspective of current LMMs' use cases, users commonly provide LMMs with screenshots that include both Japanese text and images.
Therefore, to foster core human cognitive skills and support a wide range of real-world use cases, it is essential to evaluate LMMs on sufficiently complex tasks where both the question image and the question text are presented through a visual modality.
Among the English benchmarks, MMMU-Pro~\citep{yue2024mmmupro} extends MMMU by constructing a benchmark in which both the question text and the question image are embedded within a single image, thereby enabling the evaluation of this dimension. However, no benchmark in Japanese supports such evaluation. Therefore, developing a Japanese benchmark that enables the evaluation of this dimension is essential.

In this paper, we propose \textbf{JMMMU-Pro}, an image-based Japanese Multi-discipline Multimodal Understanding benchmark. JMMMU-Pro follows the evolution from MMMU to MMMU-Pro and is constructed by embedding each of the 1,320 question texts and question images from the original JMMMU tasks into a single composite image.
Built on top of the established JMMMU, it enables an apple-to-apple comparison between JMMMU-Pro and JMMMU, which provides a meaningful signal of a model's visual cognitive abilities. Consequently, JMMMU-Pro offers both high usability and highly informative feedback for model developers.

For the construction of JMMMU-Pro, we propose a new benchmark creation methodology called \textbf{Vibe Benchmark Construction}. In this framework, an image generation model plays the primary role in producing the visual question, while humans simply check the outputs and, when necessary, refine the prompts before regenerating the images, thereby ensuring consistent quality.
Previously, when creating image-based benchmarks (\eg~MMMU-Pro), all questions had to be created manually, which was not scalable and resulted in substantial human cost. 
In contrast, Vibe Benchmark Construction leverages Nano Banana Pro~\citep{gemini3pro_image_2025}, a state-of-the-art image generation model with exceptional photorealism. Nano Banana Pro can not only generate highly realistic images but also accurately embed Japanese text within them. By supplying both the question image and the question text and prompting the model to integrate them into a single composite image, we can generate image questions with a diverse set of backgrounds and layouts.

Our Vibe Benchmark Construction approach offers several advantages over manual construction: (i) it is highly scalable, (ii) it requires minimal human effort, and (iii) it enables controllable generation of diverse layouts. In particular, given the rapid progress of image generation models, we expect Vibe Benchmark Construction to serve as an increasingly effective guideline for future benchmark development of image-based benchmarks. Through generation with Nano Banana Pro and subsequent human checking, approximately 95\% of all the questions in JMMMU-Pro were generated by Nano Banana Pro. 

In our experiments, we evaluated a total of 14 LMMs, including representative closed-source LMMs (\ie~GPT-5.2~\citep{gpt_5_2_2025} and Gemini3Pro~\citep{gemini_models_2025}), English-centric open-source LMMs (\eg~LLaVA-OneVision-1.5~\citep{an2025llava}), multilingual open-source LMMs (\eg~Qwen3VL-8B~\citep{Bai2025Qwen3VLTR} and AyaVision~\citep{dash2025aya}), and Japanese open-source LMMs (\eg~Sarashina2.2-Vision-3B~\citep{sbintuitions_sarashina2.2_vision_3b} and Heron-NVILA-Lite~\citep{turingmotors_heron_nvila_lite_15b}).
Our key experimental findings are summarized as follows:

\begin{enumerate}
\item Open-source LMMs struggle substantially on JMMMU-Pro. Even the best-performing open-source model scores below 50\%, and many LMMs achieve performance close to random guessing.
\item When compared with JMMMU, most LMMs show a drop in performance on JMMMU-Pro. In particular, open-source LMMs exhibit a decrease ranging up to 23\%.
\item Recent strong reasoning-based closed-source LMMs perform considerably well on JMMMU-Pro, revealing a substantial and concerning gap between closed-source and open-source LMMs.
\item Through detailed analysis, we find that although a major source of failure is the lack of Japanese OCR ability, strong OCR alone is not sufficient to solve JMMMU-Pro. This suggests that solving JMMMU-pro requires improving both OCR capability and the ability to interpret language and vision in an integrated manner through visual perception.
\end{enumerate}

Our contributions are summarized as follows:
\begin{itemize}
\item \textbf{Construction of JMMMU-Pro:}
We extend JMMMU and introduce JMMMU-Pro, a benchmark that embeds each question image and its corresponding text into a single image, enabling the evaluation of integrated visual-textual understanding through visual perception.
\item \textbf{Proposal of Vibe Benchmark Construction:}  
We propose Vibe Benchmark Construction, a dataset creation framework in which a powerful generation model drives the construction process, while humans only perform checking and minor prompt adjustments. With the continued progress of image generation models, we expect this approach to serve as an efficient and scalable guideline for future benchmark development.
\item \textbf{Encouraging Future Efforts in the Open-Source Community:}  
Our results show that open-source LMMs struggle heavily on JMMMU-Pro, highlighting a substantial gap with closed-source LMMs. JMMMU-Pro provides a valuable benchmark that can motivate and guide the open-source community in closing this gap.
\end{itemize}

\section{Related Work}
\label{sec:related_work}
\textbf{Large Multimodal Models (LMMs).}
Following the success of large language models (LLMs), many LMMs have been developed with improved knowledge and instruction-following capabilities~\citep{liu2023visual, liu2023improved, liu2024llavanext, li2024llava, ye2023mplug2, zhao2023svit, li2023otter, monajatipoor2023metavl, zhao2023mmicl}. With recent advances in multilingual LLMs~\citep{team2024qwen2, coherelabs_c4ai-command-r7b-12-2024}, both English-centric LMMs with multilingual capabilities~\citep{li2024llava, an2025llava, tong2024cambrian} and fully multilingual LMMs~\citep{yue2024pangea, dash2025aya, microsoft_phi4_multimodal_instruct_2025} have emerged.
In parallel, several LMMs specialized for Japanese have also been developed~\citep{sbintuitions_sarashina2_vision_14b, sbintuitions_sarashina2.2_vision_3b, baek2025harnessing, sasagawa-etal-2025-constructing, turingmotors_heron_nvila_lite_15b}.
However, these models have not been evaluated on tasks that require solving Japanese questions that rely on integrated visual-textual understanding through visual perception.
This highlights the need for a dedicated benchmark that can systematically evaluate such integrated visual-textual understanding capabilities in Japanese.

\textbf{LMM Benchmarks.}
Among various recent benchmarks~\citep{li2023seed, liu2023visual, liu2023mmbench, lu2023mathvista, yue2024mmmu, miyai2024unsolvable}, MMMU~\citep{yue2024mmmu} has become the most widely used benchmark for assessing progress in LMMs. MMMU requires advanced, university-level knowledge and reasoning across a wide range of disciplines, enabling comprehensive and expert-level evaluation.
More recently, MMMU-Pro~\citep{yue2024mmmupro} extends this evaluation paradigm by embedding both the question text and the question image into a single image, challenging models to truly ``see'' and ``read'' simultaneously, mirroring how humans naturally process complex scenes in which text and visuals are interleaved. Unlike traditional OCR-related benchmarks~\citep{singh2019towards, liu2024ocrbench} or DocVQA~\citep{mathew2021docvqa}, MMMU-Pro requires not only text recognition but also complex reasoning that integrates both visual and textual information, thereby pushing the capabilities of LMMs beyond standard document understanding. As a result, MMMU-Pro has been widely adopted in the development of recent state-of-the-art LMMs~\citep{an2025llava, openai_gpt5_2025, deepmind_gemini_2025, abouelenin2025phi}.
However, MMMU-Pro evaluates only English, leaving abilities in other languages unassessed. Therefore, developing JMMMU-Pro to evaluate the ability to understand images and text in a unified manner in Japanese is an important next step.

\textbf{Japanese LMM Benchmarks.}
The development of Japanese LMM benchmarks remains behind that of English benchmarks. Many existing studies focus primarily on common-sense knowledge and do not adequately cover expert-level domains~\citep{javgvqa,llavabenchja,llavabenchinthewild-JA,inoue2024heron,javlmbenchinthewild,ja-multivqa}, despite the rapid progress of LMMs and the importance of evaluating such capabilities.
To address these issues, JMMMU~\citep{onohara2025jmmmu} was introduced, significantly advancing the landscape of Japanese LMM evaluation. However, JMMMU does not include tasks that require models to interpret both text embedded within images, unlike MMMU-Pro.
Benchmarks that include Japanese text within images, such as JDocQA~\citep{onami2024jdocqa} and MangaVQA~\citep{baek2025mangavqa}, do exist, but they do not require the level of complex reasoning demanded by MMMU-Pro and are therefore insufficient for driving further advances in LMMs. To address this gap, we build upon JMMMU to create JMMMU-Pro, which evaluates a model's ability to jointly understand different modalities in a more integrated manner and to perform high-level reasoning on such tasks.

\textbf{QA Construction with Generative Models.}
In the context of LLM-based and LMM-based QA construction, it is common for humans to manually edit model-generated QA pairs. As a result, it is uncommon to rely solely on iterative prompt adjustments to construct QA data. While several works adopt iterative prompt adjustments in order to eliminate human-induced prompt bias~\citep{shah2024prompt, he2025prompting}, these approaches differ fundamentally from our goal, in that our objective is to scalably produce high-quality QA data.
In the context of image-generation models, recent works have leveraged powerful image generation models to create the images of VQA benchmarks~\citep{wang2024vlbiasbench, xiao2025genderbias}.
However, these approaches still require substantial additional effort to construct the question texts through manual creation or LMM-based question generation.
The most similar work to our concept is LogicOCR~\citep{ye2025logicocr}, which uses GPT-Image-1~\citep{openai_image_generation_api} to embed English question text into images with varied layouts. LogicOCR performs manual verification and discards a portion of the generated samples. Although the approach faces limitations, such as reduced photorealism due to the capabilities of GPT-Image-1 and dataset shrinkage caused by sample filtering, it demonstrates a promising direction of leveraging modern image generation models.
We build upon this line of work and we formalize a pipeline, Vibe Benchmark Construction, in which the process of generating VQA embedded in images is primarily driven by image generation models, while humans perform verification and, when necessary, adjust prompts and regenerate the images. By defining this pipeline clearly, we provide an effective guideline for the scalable construction of future image-based VQA benchmarks.

\section{JMMMU-Pro Benchmark}
\begin{figure*}[t]
\centering
    \includegraphics[width=0.99\linewidth]{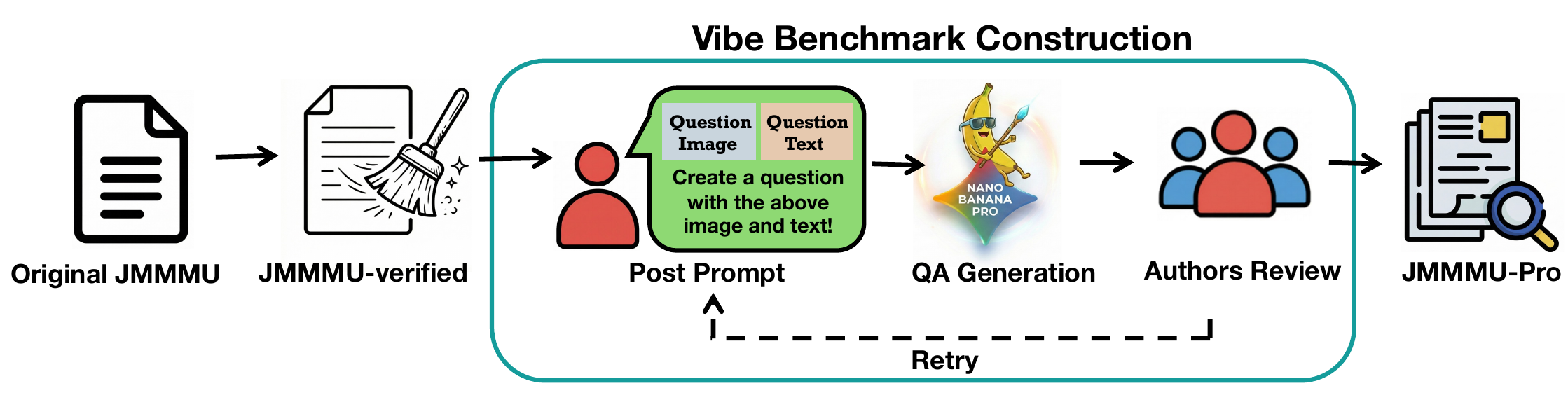}\\
   \caption{JMMMU-Pro Construction Pipeline.}
    \label{fig:pipeline}
\end{figure*}

JMMMU-Pro consists of 1,320 questions whose contents are identical to those in the original JMMMU.
\Cref{fig:pipeline} illustrates the construction pipeline. We first describe the original JMMMU and then present the core component of our approach: Vibe Benchmark Construction.

\subsection{Revisiting the JMMMU Benchmark}
JMMMU~\citep{onohara2025jmmmu} consists of 1,320 questions and 1,118 images spanning 28 subjects.
The benchmark is intentionally divided into two categories: culture-agnostic and culture-specific subjects.
The culture-agnostic subset includes 24 subjects with 720 questions, across five disciplines: (1) Art \&
Psychology, (2) Business, (3) Health \& Medicine,
(4) Science, and (5) Tech \& Engineering. The culture-specific subset comprises 600 questions across four subjects: (1) Japanese Art, (2) Japanese Heritage, (3) Japanese History, and (4) World History. 

To simplify the interpretation of JMMMU evaluation results, we converted all 50 open-ended questions into multiple-choice questions and revised two samples. We refer to this updated version as \texttt{JMMMU-verified-2025-12} (see details in \Cref{sec_appendix:verified_jmmmu}). All JMMMU scores reported in this paper are based on \texttt{JMMMU-verified-2025-12}.

\begin{figure}[t!]
  \centering
  \begin{subfigure}[b]{0.48\textwidth}
    \centering
    \includegraphics[width=\textwidth,keepaspectratio]{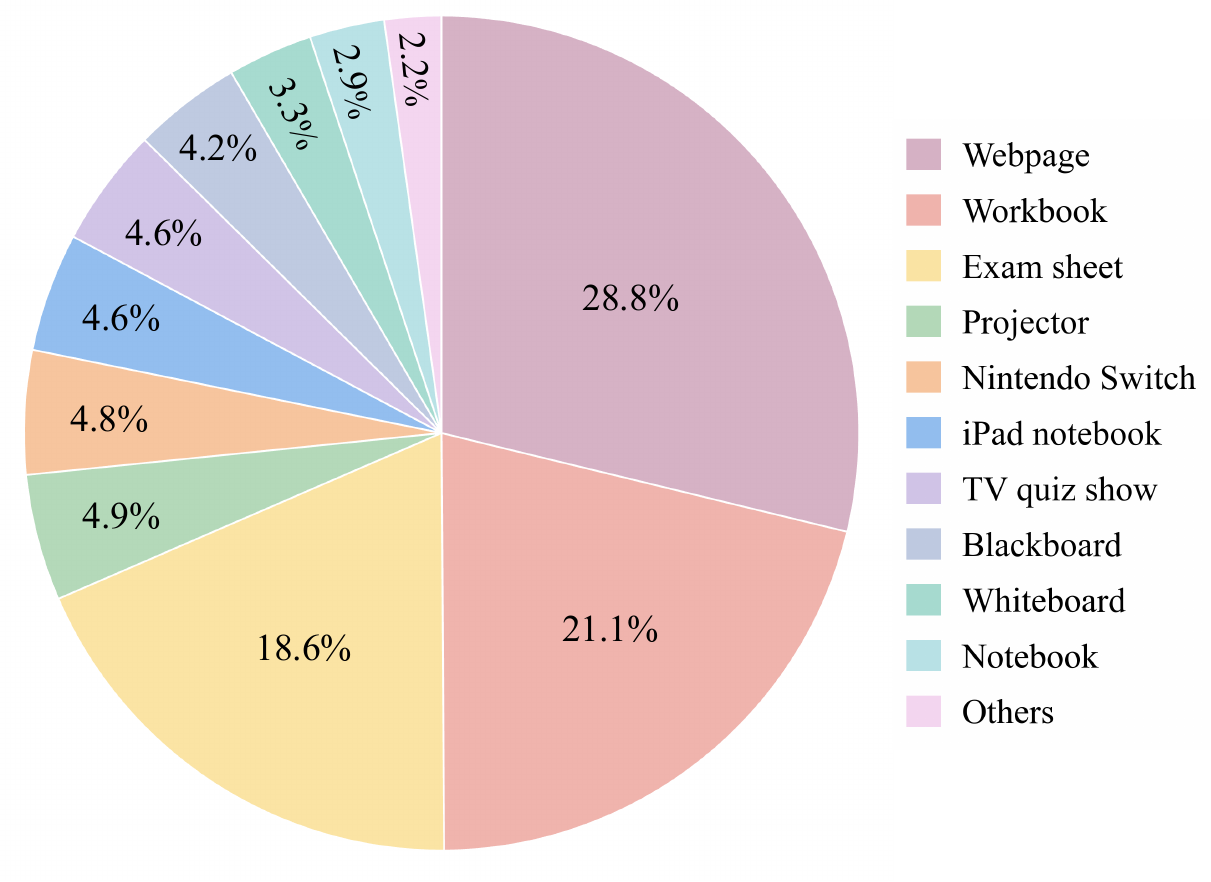}
    \caption{Background distribution}
    \label{fig:background-distribution}
  \end{subfigure}
  \hfill
  \begin{subfigure}[b]{0.48\textwidth}
    \centering
    \includegraphics[width=\textwidth,keepaspectratio]{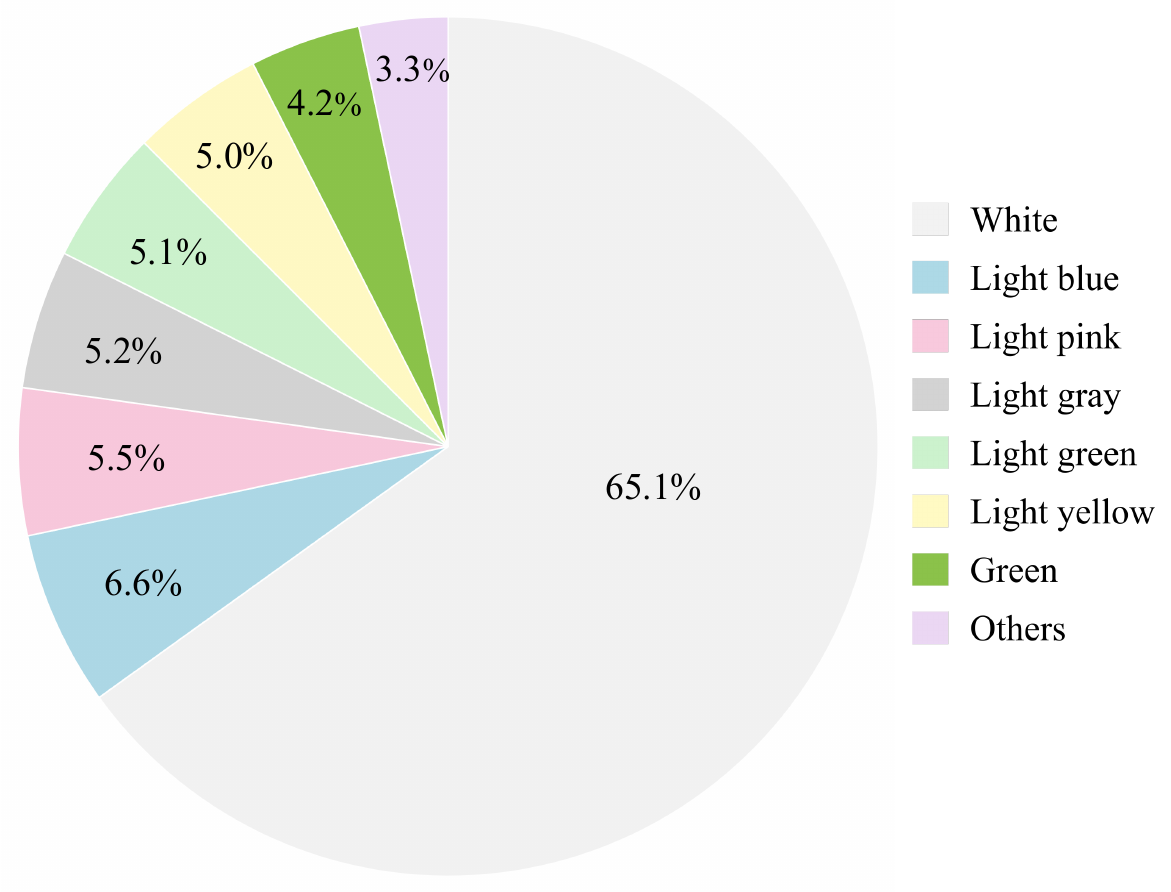}
    \caption{Color distribution}
    \label{fig:color-distribution}
  \end{subfigure}
  \caption{Distribution of background and its color in the JMMMU-Pro benchmark.}
  \label{fig:distributions}
\end{figure}

\begin{figure*}[h]
\centering
    \includegraphics[width=0.99\linewidth]{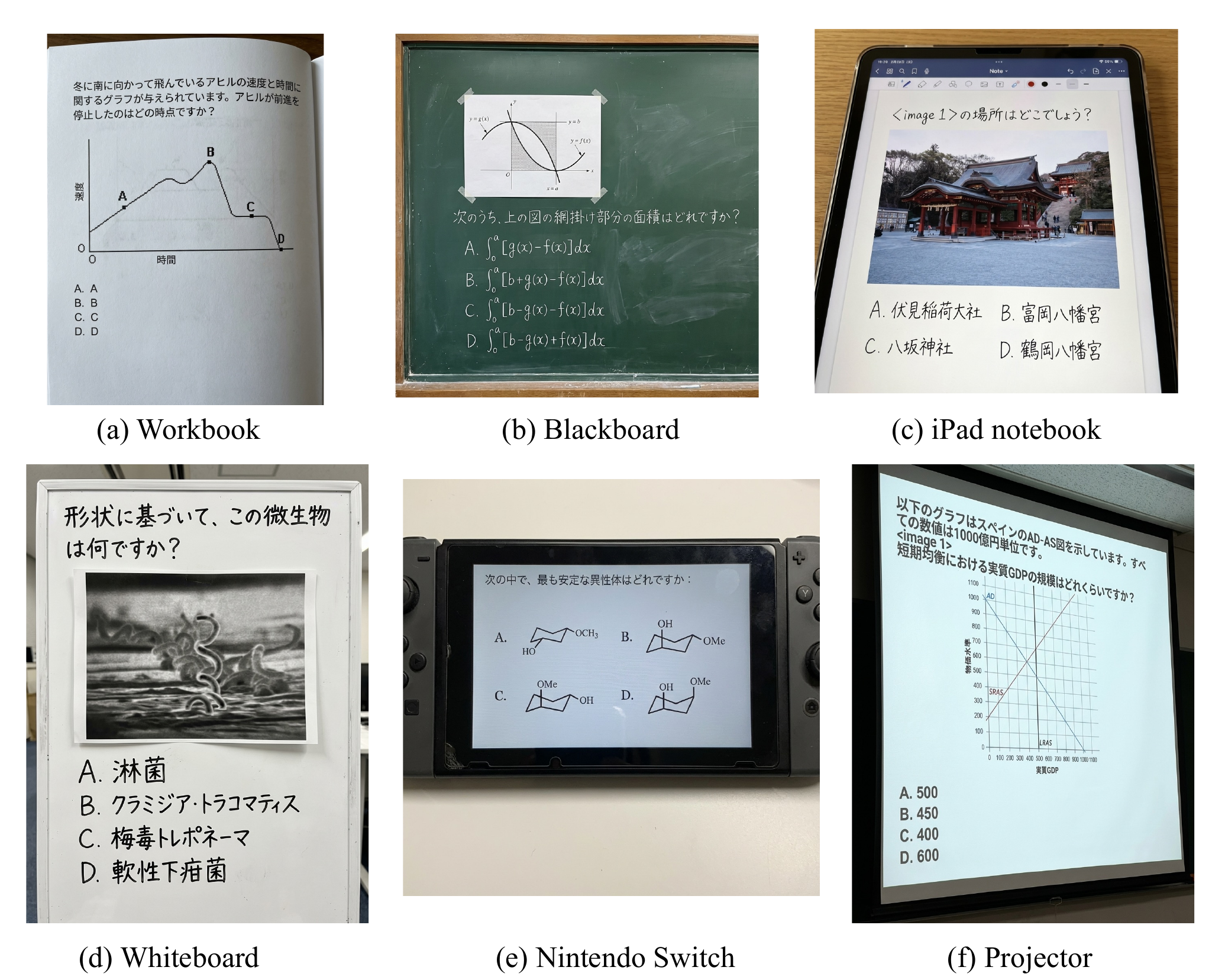}\\
   \caption{JMMMU-Pro samples.}
    \label{fig:samples}
\end{figure*}

\subsection{Definition of Vibe Benchmark Construction}
Vibe Benchmark Construction is a methodology in which an image generation model plays the primary role in producing the VQA problem images, while humans only verify the outputs and adjust the prompts when necessary to ensure quality.
Although previous VQA benchmarks have used synthetic images generated by image generation models, these models have played only a supplementary role, producing just the visual part while the question text still had to be created separately by humans or by an LMM, incurring additional cost.

In contrast, the key distinction of our proposed Vibe Benchmark Construction is that the VQA creation process is carried out by the image generation model, with humans intervening solely for verification and prompt refinement.
This paradigm is particularly effective for image-based VQA, where humans cannot easily edit content directly inside the image in the same way as text-based QA. By letting the model handle generation and restricting human effort to \textit{adjusting the prompt until a satisfactory image is produced}, the method enables efficient and scalable construction of benchmarks, especially in domains like image-based VQA, where dataset creation is difficult. A more detailed comparison with existing work is provided in \Cref{sec:related_work}.

\subsection{Detailed Pipeline of Vibe Benchmark Construction}
For image generation, we use Nano Banana Pro via its API interface (\texttt{gemini-3-pro-image-preview}). The image resolution is set to 1K.
Below, we describe our prompt design process and the workflow for human checking and regeneration.

\textbf{Prompt Selection and Image Generation.}
We first selected a prompt template through preliminary experiments. Specifically, we used the prompt template in \Cref{sec_appendix:prompt} and varied the following six components as parameters to generate a diverse set of images (shown in \Cref{fig:samples}):
\textit{1. Background}: chosen from workbook, exam sheet, whiteboard, blackboard, projector, iPad notebook, webpage, Nintendo Switch, and TV quiz show.
\textit{2. Background Color}: selected from white, light green, light yellow, light pink, light gray, and light blue. Note that certain backgrounds have a fixed color (\eg~a whiteboard is always white), and we account for such constraints.
\textit{3. Font}: chosen from handwritten text, computer font, thick computer font, thin computer font, and manga-style computer font.
\textit{4. Margin}: selected as either small or large.
\textit{5. State}: chosen from photo by smartphone, screenshot by PC, and screenshot by smartphone. 
\textit{6. Aspect Ratio}: selected from 9:16, 16:9, 3:4, and 1:1.
We show the statistics for the two most controllable factors: \textit{Background} and \textit{Background Color} in \Cref{fig:distributions}.

In addition, while JMMMU includes image tags such as <image 1> within the question text, we found that Nano Banana Pro does not allow control over these image tags. To address this issue, we attempted to include explicit instructions in the prompt (\eg~``keep the image tag in the question''). However, such instructions significantly degraded the quality of the generated images. We hypothesize that this occurs because Nano Banana Pro internally uses similar image-tag tokens, and explicit instructions about them may interfere with its generation process. Therefore, we intentionally avoid giving any special instructions regarding image tags in the prompt.

\textbf{Human Checking and Regeneration.}
We performed author reviews of the generated images with a custom-built annotation tool.
In these reviews, we checked that the generated text and images matched the originals exactly. As mentioned above, controlling image tags in the question text is difficult, so we allowed variations in the tags as long as the generated item remained a valid question.

In the first review round, 71\% of the questions passed. The remaining 29\% failed primarily because the question image had been replaced with an unrelated image, the text within the image was unreadable, parts of the question text were missing or incorrect, or the generated image was visually unnatural. These examples are shown in \Cref{fig:failure_banana}. For these failed cases, we regenerated the images with the same prompt or prompts with minor prompt adjustments.

After completing the full set of VQA questions across several rounds, we performed a final cross-check to eliminate inconsistencies in the authors' evaluation standards.

\textbf{Manual Construction.}
We manually created 67 samples that Nano Banana Pro had difficulty generating. These cases had the following characteristics:
1. long question text (16 samples), 2. small or difficult-to-render text within the question image (36 samples), 3. extreme aspect ratios (2 samples), 4. domains that are inherently difficult to generate, such as chemical formulas or musical notation (8 samples), and 5. cases rejected by Nano Banana Pro due to policy constraints (5 samples).
These examples are shown in \Cref{fig:human_created}.

\section{Experiments}
\subsection{Setup}
\begin{table*}[t]
\vspace{-3mm}
\centering
\centering
\begin{tabular}{@{}lc|c|cc|cc@{}}
\toprule
Model & JMMMU-Pro & \color{gray}JMMMU & CS Pro & \color{gray}CS & CA Pro & \color{gray}CA \\
 & (1320) & \color{gray}(1320) & (600) & \color{gray}(600) & (720) & \color{gray}(720) \\
\midrule
\textbf{Random} & & & & & &  \\
~~~~Random Choice & 27.05 & \color{gray}27.05 & 26.33 &  \color{gray}26.33 & 27.64  & \color{gray}27.64 \\
~~~~Frequent Choice  & 27.73 & \color{gray}27.73 & 25.33 &  \color{gray}25.33 & 29.72 & \color{gray}29.72 \\
\midrule
\textbf{Multilingual Open LMMs} & & & & & &  \\
~~~~Qwen3-VL-8B & 47.27 & \color{gray}52.88 & 47.50 & \color{gray}55.83 & 47.08 & \color{gray}50.42 \\
~~~~Qwen2.5-VL-7B & 45.00 & \color{gray}47.65 & 46.67 & \color{gray}54.00 & 43.61 & \color{gray}42.36 \\
~~~~Phi-4-multimodal & 31.82 & \color{gray}39.55 & 28.83 & \color{gray}38.00 & 34.31 & \color{gray}40.83 \\
~~~~Aya-Vision-8B & 26.74 & \color{gray}37.73 & 27.00 & \color{gray}40.33 & 26.53 & \color{gray}35.56 \\
~~~~Pangea-7B & 23.41 & \color{gray}37.50 & 21.67 & \color{gray}47.17 & 24.86 & \color{gray}29.44 \\
\midrule
\textbf{English-centric Open LMMs} & & & & & &  \\
~~~~LLaVA-OV-1.5-8B & 31.97 & \color{gray}51.74 & 28.00 & \color{gray}53.33 & 35.28 & \color{gray}50.42 \\
~~~~LLaVA-OV-7B & 27.35 & \color{gray}41.14 & 26.50 & \color{gray}43.83 & 28.06 & \color{gray}38.89 \\
~~~~InternVL2.5-8B & 31.21 & \color{gray}41.36 & 29.00 & \color{gray}43.33 & 33.06 & \color{gray}39.72 \\
\midrule
\textbf{Japanese Open LMMs} & & & & & &  \\
~~~~Sarashina2.2-V-3B & 42.88 & \color{gray}47.95 & 54.00 & \color{gray}61.50 & 33.61 & \color{gray}36.67 \\
~~~~Sarashina2-V-14B & 30.68 & \color{gray}37.27 & 32.33 & \color{gray}43.17 & 29.31 & \color{gray}32.36 \\
~~~~Sarashina2-V-8B & 27.88 & \color{gray}39.62 & 27.00 & \color{gray}51.00 & 28.61 & \color{gray}30.14 \\
~~~~Heron-NVILA-Lite-15B & 26.97 & \color{gray}50.15 & 26.67 & \color{gray}59.17 & 27.22 & \color{gray}42.64 \\
\midrule
\textbf{Closed LMMs} & & & & & &  \\
~~~~Gemini3Pro (reasoning high) & 87.04 & \color{gray}89.77 & 95.00 & \color{gray}95.00 & 80.42 & \color{gray}85.42 \\
~~~~GPT-5.2 (reasoning high) & 83.33 & \color{gray}84.47 & 88.33 & \color{gray}85.50 & 79.17 & \color{gray}83.61 \\
\bottomrule
\end{tabular}
\caption{\textbf{Main Results on JMMMU-Pro and JMMMU.} Overall, most open-source LMMs show substantial performance degradation on JMMMU-Pro compared to JMMMU, while closed-source LMMs maintain strong performance, highlighting a significant gap in integrated Japanese visual-textual understanding.}
\label{table:main_results}
\end{table*}
\textbf{Baseline LMMs.}
For a more comprehensive evaluation, we assess a diverse set of state-of-the-art LMMs. In particular, for open-source models, we select representative models from three categories, English-centric LMMs, multilingual LMMs, and Japanese LMMs, to ensure that our evaluation accurately captures current progress in each subfield.
We mainly use LMMs-Eval~\cite{zhang2024lmms} for our experiments. We set the \texttt{temperature} to 0 to open-source LMMs (the default setting for closed-source LMMs), and set \texttt{max\_tokens} to be configured to be long enough so that the response would not be cut off.

\textit{Closed-source LMMs}: GPT-5.2~\citep{gpt_5_2_2025}, Gemini3Pro~\citep{gemini_models_2025}.

\textit{English-Centric Open-source LMMs}:
LLaVA-OneVision-7B~\citep{li2024llava},
LLaVA-OneVision-1.5-8B~\citep{an2025llava},
InternVL2.5-8B~\citep{chen2024expanding},
InternVL3-14B~\citep{zhu2025internvl3},

\textit{Multilingual Open-source LMMs}:
Qwen2.5VL-7B~\citep{bai2025qwen2}, Qwen3VL-8B~\citep{Bai2025Qwen3VLTR},
Phi-4Multimodal~\citep{microsoft_phi4_multimodal_instruct_2025},
Pangea-7B~\citep{yue2024pangea}, Aya Vision-8B~\citep{dash2025aya}

\textit{Japanese Open-source LMMs}: 
Sarashina2-Vision-8B and 14B~\citep{sbintuitions_sarashina2_vision_14b}, Sarashina2.2-Vision-3B~\citep{sbintuitions_sarashina2.2_vision_3b}, Heron-NVILA-Lite-15B~\citep{turingmotors_heron_nvila_lite_15b}

\textbf{Inference Prompt.}
The inference prompt is based on the setup in JMMMU~\citep{onohara2025jmmmu} and MMMU-Pro~\citep{yue2024mmmupro}. 
Following MMMU-Pro~\citep{yue2024mmmupro}, we evaluate the open-source LMMs with both Direct and CoT prompts (as shown in \Cref{sec_appendix:prompt}), and report the higher ones in the overall results. For the closed-source LMMs, they perform reasoning regardless of the prompt types, so we evaluated them using only the Direct Prompt.
Full results are shown in \Cref{sec_appendix:full_results}.

\subsection{Main Results}
We present the experimental results in \Cref{table:main_results}. The key findings from these results are as follows.

\textbf{F1. All open-source LMMs struggle significantly on JMMMU-Pro.}
As shown in \Cref{table:main_results}, open-source LMMs perform poorly on JMMMU-Pro, with the best model, Qwen3-VL-8B, achieving only 45.83, indicating substantial room for improvement. Furthermore, nine models perform less than 32\%, close to random guessing. 
These results highlight that JMMMU-Pro poses a challenging and valuable benchmark for evaluating and advancing open-source LMMs.

\textbf{F2. Most open-source LMMs exhibit a significant performance drop compared to JMMMU.}
As shown in \Cref{table:main_results}, most open-source LMMs, except for Qwen2.5-VL-7B, show a substantial decline in accuracy on JMMMU-Pro relative to JMMMU.
Moreover, when we compare the CS and CA subsets, we find that models with a clear performance gap between the JMMMU's two subsets are similarly low on both in JMMMU-Pro. This suggests that their weakness lies in a fundamental lack of vision-side understanding, rather than in the type of question.
These results demonstrate that JMMMU-Pro provides valuable feedback to model developers when used in comparison with JMMMU.

\textbf{F3. Closed-source LMMs achieve substantially higher performance on JMMMU-Pro, revealing a serious gap relative to open-source models.}
As shown in \Cref{table:main_results}, closed-source LMMs obtain notably high scores on JMMMU-Pro. This indicates that these models already possess the ability to seamlessly integrate visual and textual information and interpret them through visual perception.
Importantly, the strong performance of closed-source models does not diminish the value of JMMMU-Pro. Instead, it highlights the crucial role of JMMMU-Pro as a benchmark for guiding the development of open-source LMMs. Given the considerable performance gap between closed-source and open-source LMMs, reducing this gap is an essential goal for the community.

\section{Analysis}
\subsection{Impact of CoT Prompting}
\begin{figure*}[t]
\centering
    \includegraphics[width=0.99\linewidth]{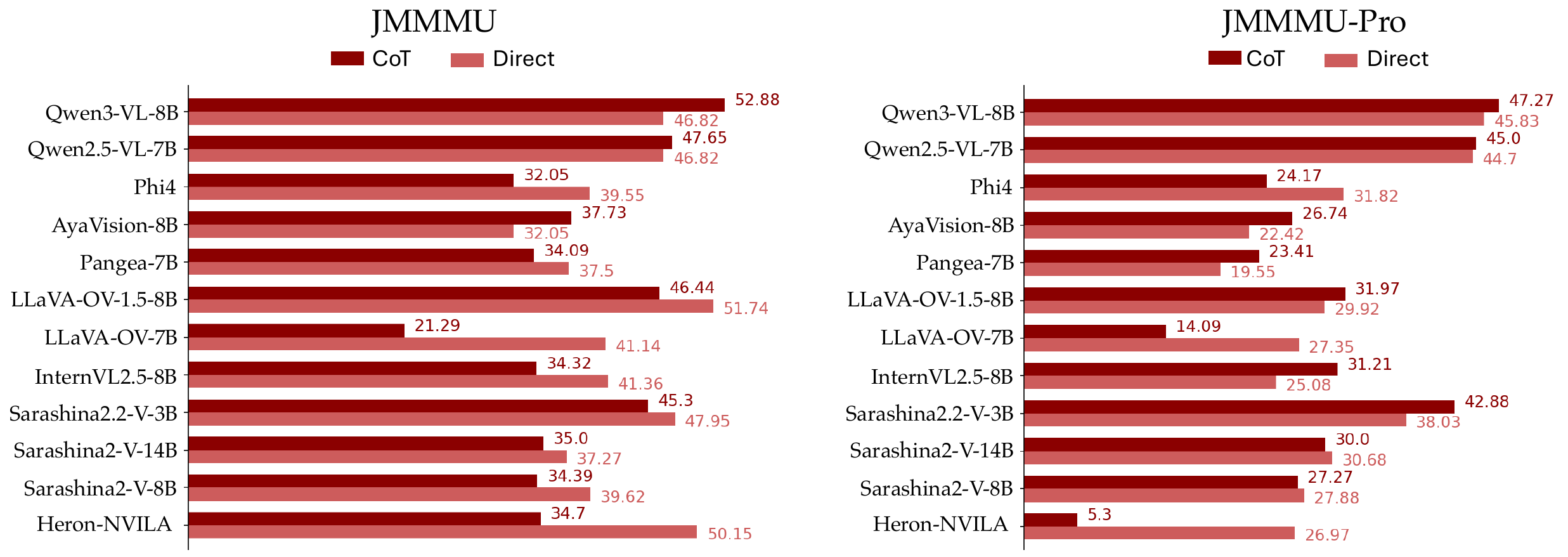}\\
   \caption{Impact of CoT prompting in JMMMU and JMMMU-Pro.}
    \label{fig:cot_analysis}
\end{figure*}

We examine the effectiveness of Chain-of-Thought (CoT) prompting on JMMMU-Pro and JMMMU. The results are shown in \Cref{fig:cot_analysis}. These results indicate that the effectiveness of CoT varies depending on the model and the evaluation setting for both JMMMU-Pro and JMMMU.
For example, on JMMMU-Pro, 7 out of the 12 LMMs achieve higher performance with CoT prompting, whereas on JMMMU, only 3 models benefit from CoT. Moreover, when examined on a per-model basis, LMMs such as Pangea-7B, LLaVA-OV-1.5-8B, InternVL2.5-8B, and Sarashina2.2-V-3B show different prompt preferences between JMMMU and JMMMU-Pro.
These findings suggest that optimal prompting strategies must be tailored to each model and each task, rather than relying on a single prompting approach across settings.

\subsection{Correlation with OCR Performance}
We hypothesize that the primary cause of performance degradation on JMMMU-Pro is the inability of current LMMs to perform Japanese Optical Character Recognition (OCR). To examine this hypothesis, we compute the correlation between OCR performance and JMMMU-Pro accuracy across several LMMs.

\begin{wrapfigure}[15]{r}{0.5\textwidth}
\centering
    \includegraphics[width=0.80\linewidth]{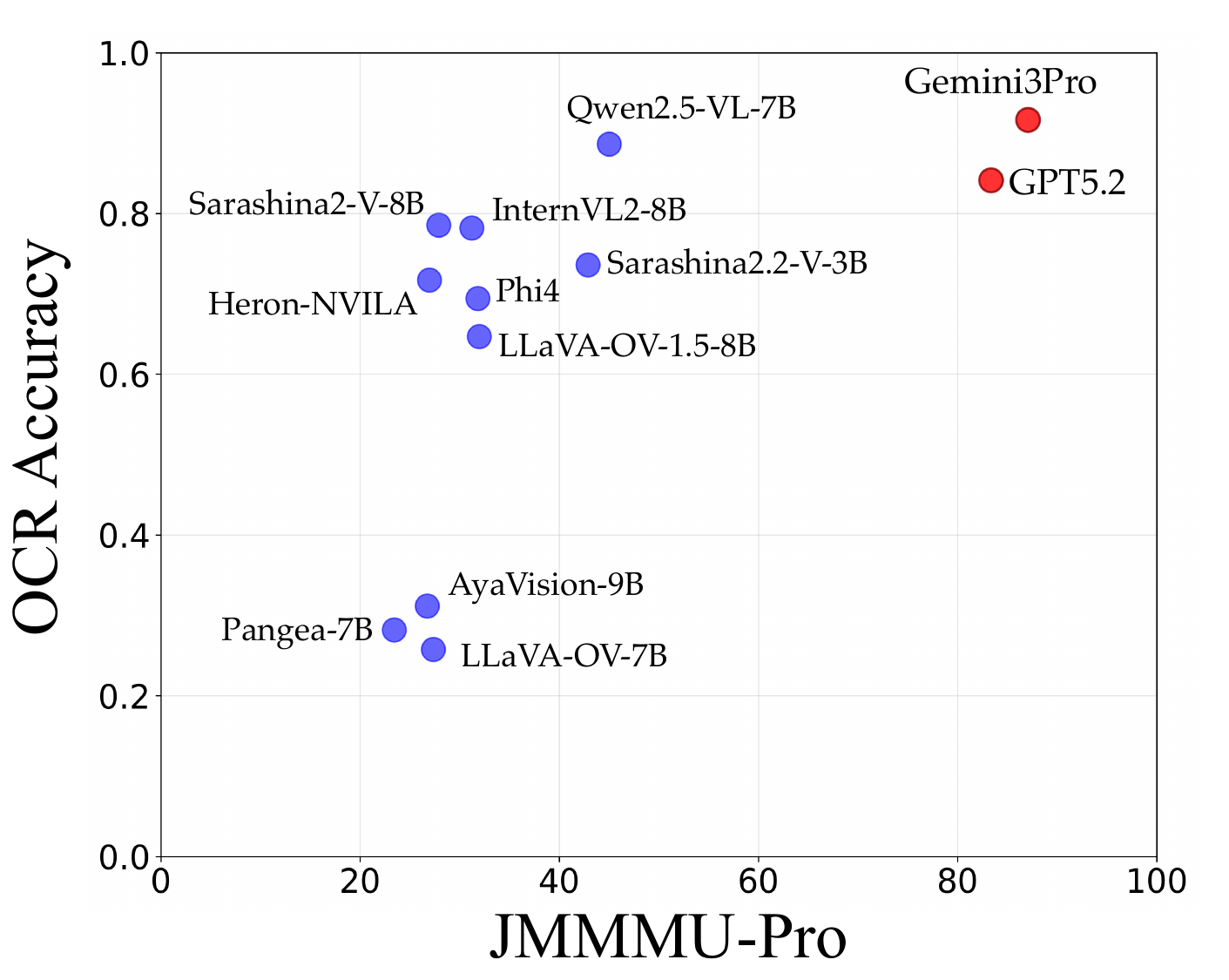}
    \captionsetup{width=0.8\linewidth}
    \caption{Correlation between OCR accuracy and JMMMU-Pro performance.}
    \label{fig:ocr_analysis}
\end{wrapfigure}

Following the evaluation setting of MMMU-Pro, we ask each LMM to extract the full text of the question and all answer choices, excluding any text from associated images. OCR accuracy is then calculated by comparing the extracted text with the original text using the Levenshtein distance, which measures the edit distance between two strings. The similarity between the extracted and original text is computed as:

\begin{equation}
\text{OCR Accuracy} = 1 - \frac{\text{Levenshtein}(\text{text}_1, \text{text}_2)}{\max(\text{len}(\text{text}_1), \text{len}(\text{text}_2))}
\end{equation}

The results are shown in \Cref{fig:ocr_analysis}. The correlation coefficient between OCR accuracy and JMMMU-Pro accuracy is 0.593. As illustrated in the figure, there is indeed a positive correlation between the two. However, high OCR ability does not necessarily translate directly into high JMMMU-Pro accuracy. For example, while Heron-NVILA and Sarashina2.2-V are comparable for OCR performance, the performance for JMMMU-Pro differs a lot.
This indicates that solving JMMMU-Pro demands not only strong OCR capabilities, but also the ability to interpret and reason over language and vision in an integrated manner through visual perception.

\subsection{Failure Examples}
We present representative failure cases of the state-of-the-art open-source LMM, Qwen3-VL-8B, in \Cref{fig:failure_1} and \Cref{fig:failure_2}. These examples demonstrate that the model makes both reasoning and perceptual errors specifically on JMMMU-Pro, suggesting that JMMMU-Pro demands deeper, more integrated visual-textual understanding that goes beyond simple OCR capabilities.
\section{Conclusion}
This paper introduces JMMMU-Pro, an image-based Japanese Multi-discipline Multimodal Understanding Benchmark, along with Vibe Benchmark Construction, a scalable methodology for creating such datasets. Our experiments show that all open-source LMMs face significant difficulty on JMMMU-Pro, highlighting its importance as a benchmark that can inspire future progress in the open-source community. We believe that JMMMU-Pro enables more rigorous evaluation of Japanese multimodal capabilities, and that Vibe Benchmark Construction provides an effective framework for the scalable development of future image-based VQA benchmarks.
\section*{Acknowledgment}
This work was partially supported by JSPS 25H01164.

\bibliographystyle{plainnat}
\bibliography{egbib.bib}

\clearpage
\newcommand\beginsupplement{%
        \setcounter{table}{0}
        \renewcommand{\thetable}{\Alph{table}}%
        \setcounter{figure}{0}
        \renewcommand{\thefigure}{\Alph{figure}}%
     }
\beginsupplement
\appendix
\section*{Appendix}
In this appendix, we describe failure cases of image generation in \Cref{sec_appendix:failure_banana}, the verified JMMMU in \Cref{sec_appendix:verified_jmmmu}, detailed prompt examples in \Cref{sec_appendix:prompt}, the full results in \Cref{sec_appendix:full_results}, and failure cases of LMM inference in \Cref{sec_appendix:failure_cases}.

\section{Image Generation Failures and Manual Construction Examples}
\label{sec_appendix:failure_banana}
\subsection{Image Generation Failures in Nano Banana Pro}
\begin{figure*}[h]
\centering
    \includegraphics[width=0.99\linewidth]{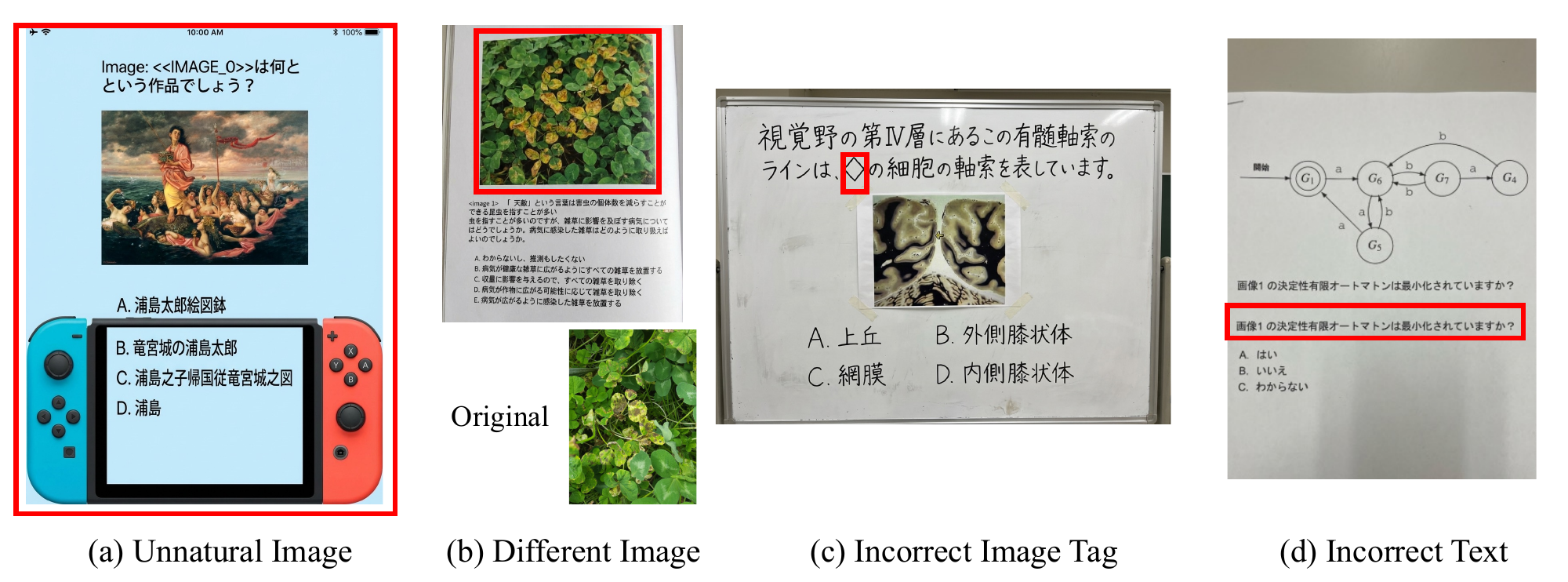}\\
   \caption{Failure examples for Nano Banana Pro.}
    \label{fig:failure_banana}
\end{figure*}

\Cref{fig:failure_banana} presents examples of image generation failures observed in Nano Banana Pro.
Nano Banana Pro can occasionally produce failures such as those shown in the figure.
We attribute these failures to the inherent diversity of outputs produced by generative models.
Therefore, it is crucial to manually inspect and filter such outputs to ensure correctness.

\subsection{Manual Construction Examples}
\begin{figure*}[h]
\centering
    \includegraphics[width=0.99\linewidth]{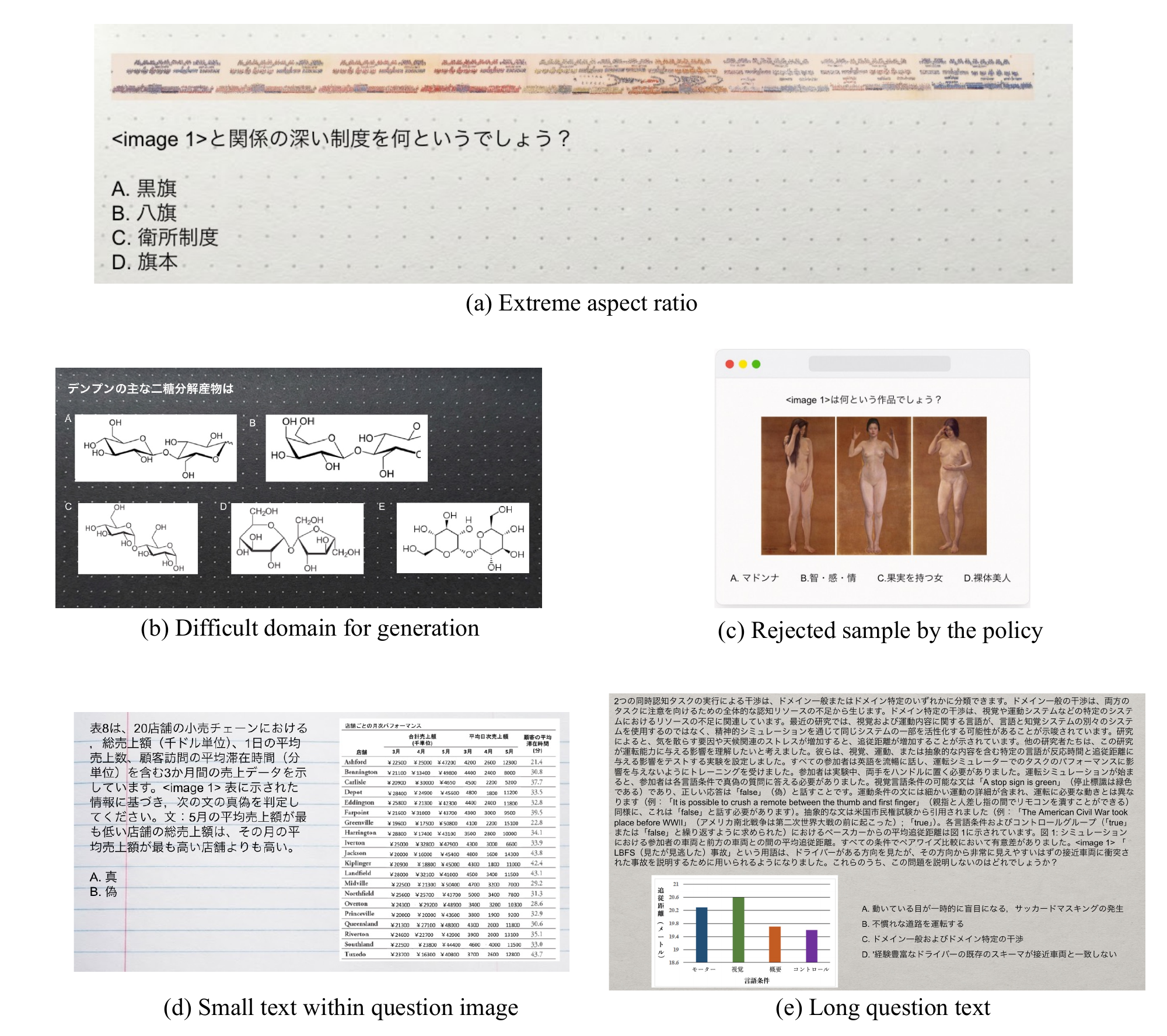}\\
   \caption{Manual construction examples.}
    \label{fig:human_created}
\end{figure*}

\Cref{fig:human_created} shows examples that were manually constructed.
Images with these characteristics were inherently difficult to generate automatically using Nano Banana Pro.
Therefore, we found that not all images can be created through the Vibe Benchmark Construction pipeline.

\section{Verifying the original JMMMU Benchmark}
\label{sec_appendix:verified_jmmmu}
We first verify the existing JMMMU benchmark by correcting its samples and refining its evaluation protocol.

\textbf{Sample Corrections.}
Although most questions in JMMMU are multiple-choice, 50 questions in the culture-agnostic subset, which translated directly from MMMU, are open-ended. Prior work~\citep{sbintuitions_sarashina2.2_vision_3b} has noted that these open-ended questions introduce additional complexity in interpreting evaluation results, and thus converted them into multiple-choice format for their experiments. Following this approach, we also convert all open-ended questions in JMMMU into multiple-choice questions. We provide the problem statement and the correct answer to an LLM (\eg~GPT-5~\citep{openai_gpt5_2025}) and instruct it to propose plausible false choices. We then manually verify the generated choices to ensure quality, correcting cases where false options might unintentionally match the correct answer due to numerical precision, ambiguity, or formatting issues.
Additionally, we identified and corrected errors in two samples (\texttt{test\_Japanese\_Art\_120}, \texttt{validation\_Agriculture\_1}), where the answer or question text contained mistakes.

\textbf{Revision of the Evaluation Procedure.}
We also revised the evaluation procedure in JMMMU. First, we found that the answer-parsing algorithm used in JMMMU often fails on recent reasoning models, whose outputs tend to be longer. In particular, when models list all options as part of their reasoning, the parser may incorrectly extract the predicted choice. To address this, we modify the parser to ignore such auxiliary option listings. 
In addition, when parsing fails, JMMMU selects an answer at random. We remove this random selection step, as it obscures whether a model has genuinely failed to produce a valid option. Our modification instead marks such cases explicitly as incorrect, allowing a clearer distinction between invalid output and an actual wrong answer.

To avoid confusion in the community, we name this corrected version of the benchmark \texttt{JMMMU-verified-2025-12}. All JMMMU scores reported in this paper are based on \texttt{JMMMU-verified-2025-12}.

\section{Prompt}
\label{sec_appendix:prompt}
\subsection{Image Generation Prompt}
Below, we present the prompts used to generate images with Nano Banana Pro. The base prompt is designed for the single-image setting, while for the multiple-image setting, the prompt is modified depending on whether images are included in the options.

\begin{tcolorbox}[colback=gray!10, colframe=gray!40!black, title=Base prompt]
    Your role is to create an image-based question. You must not derive or provide the answer.
Please create an image that looks as if the pictures, text, and options were actually placed or written on a surface, such as a notebook, a sheet of paper, a webpage, or other backgrounds, and then captured either by being photographed with a phone camera or taken as a smartphone/PC screenshot.

\vspace{6pt}
**Critical Instruction**

\vspace{1pt}
1. Insert the images exactly as they are.

\vspace{1pt}
2. Do not change the character of any text in the Image or the Question. Keep the original character exactly as it appears.

\vspace{1pt}
3. Do not derive the answer. Your task is to create an image-based question.

\vspace{1pt}
4. Make sure the pasted images blend naturally into the notebook background while still retaining a subtle "pasted" feel. For cases such as a blackboard or a notebook, keep a slight pasted effect. For things like a webpage, a projector screen, or printed material, make the pasted images blend in more naturally without an obvious pasted look, while still keeping the boundary between the Question and the Image clearly distinguishable.

\vspace{1pt}
5. Please make the font of the text within the images follow the original font in the given image as closely as possible. The font for the question and option text should follow the instructions provided below.

\vspace{1pt}
6. Paste the image without making any edits.

\vspace{6pt}
**More Detailed Conditions**

\vspace{1pt}
1. The image should reflect \{state\}.

\vspace{1pt}
2. The font used in the question and options should be \{font\}.

\vspace{1pt}
3. The background should be \{background\}.

\vspace{1pt}
4. The background color should be \{color\}.

\vspace{1pt}
5. The marginal space should be \{margin\}.

\vspace{6pt}
Image:

\vspace{1pt}
As attached

\vspace{1pt}
Question:

\vspace{1pt}
\{question\}

\vspace{1pt}
\{options\}
\end{tcolorbox}

\begin{tcolorbox}[colback=gray!10, colframe=gray!40!black, title=Prompt for multiple images (wo option images)]
    Your role is to create an image-based question. You must not derive or provide the answer.
Please create an image that looks as if the pictures, text, and options were actually placed or written on a surface, such as a notebook, a sheet of paper, a webpage, or other backgrounds, and then captured either by being photographed with a phone camera or taken as a smartphone/PC screenshot.

\vspace{6pt}
**Critical Instruction**

\vspace{1pt}
1. Insert the images exactly as they are.

\vspace{1pt}
2. Do not change the character of any text in the Image or the Question. Keep the original character exactly as it appears.

\vspace{1pt}
3. Do not derive the answer. Your task is to create an image-based question.

\vspace{1pt}
4. Make sure the pasted images blend naturally into the notebook background while still retaining a subtle "pasted" feel. For cases such as a blackboard or a notebook, keep a slight pasted effect. For things like a webpage, a projector screen, or printed material, make the pasted images blend in more naturally without an obvious pasted look, while still keeping the boundary between the Question and the Image clearly distinguishable.

\vspace{1pt}
5. Please make the font of the text within the images follow the original font in the given image as closely as possible. The font for the question and option text should follow the instructions provided below.

\vspace{1pt}
6. Paste the image without making any edits.

\vspace{1pt}
7. When there are multiple images, place them from left to right, one by one.

\vspace{6pt}
**More Detailed Conditions**

\vspace{1pt}
1. The image should reflect \{state\}.

\vspace{1pt}
2. The font used in the question and options should be \{font\}.

\vspace{1pt}
3. The background should be \{background\}.

\vspace{1pt}
4. The background color should be \{color\}.

\vspace{1pt}
5. The marginal space should be \{margin\}.

\vspace{6pt}
Image:

\vspace{1pt}
As attached

\vspace{1pt}
Question:

\vspace{1pt}
\{question\}

\vspace{1pt}
\{options\}
\end{tcolorbox}

\begin{tcolorbox}[colback=gray!10, colframe=gray!40!black, title=Prompt for image options]
    Your role is to create an image-based question. You must not derive or provide the answer.
Please create an image that looks as if the pictures, text, and options were actually placed or written on a surface, such as a notebook, a sheet of paper, a webpage, or other backgrounds, and then captured either by being photographed with a phone camera or taken as a smartphone/PC screenshot.

\vspace{6pt}
**Critical Instruction**

\vspace{1pt}
1. Insert the images exactly as they are.

\vspace{1pt}
2. Do not change the character of any text in the Image or the Question. Keep the original character exactly as it appears.

\vspace{1pt}
3. Do not derive the answer. Your task is to create an image-based question.

\vspace{1pt}
4. Make sure the pasted images blend naturally into the notebook background while still retaining a subtle “pasted” feel. For cases such as a blackboard or a notebook, keep a slight pasted effect. For things like a webpage, a projector screen, or printed material, make the pasted images blend in more naturally without an obvious pasted look, while still keeping the boundary between the Question and the Image clearly distinguishable.

\vspace{1pt}
5. Please make the font of the text within the images follow the original font in the given image as closely as possible. The font for the question and option text should follow the instructions provided below.

\vspace{1pt}
6. Paste the image without making any edits.

\vspace{1pt}
7. For the <image> tags in options (e.g., <image 1>, <image 2>), please replace each tag with the actual image. The images correspond to the tags in order, with the first image being <image 1>, the second <image 2>, and so on.

\vspace{6pt}
**More Detailed Conditions**

\vspace{1pt}
1. The image should reflect \{state\}.

\vspace{1pt}
2. The font used in the question should be \{font\}.

\vspace{1pt}
3. The background should be \{background\}.

\vspace{1pt}
4. The background color should be \{color\}.

\vspace{1pt}
5. The marginal space should be \{margin\}.

\vspace{6pt}
Image:

\vspace{1pt}
As attached

\vspace{1pt}
Question:

\vspace{1pt}
\{question\}

\vspace{1pt}
\{options\}
\end{tcolorbox}

\subsection{LMM Inference Prompt.}
Below, we present the prompt used for LMM inference.

\begin{promptbox}[Inference Prompts: Direct]{lightgreen}
\textbf{JMMMU:}\\
      ``\textit{与えられた選択肢の中から最も適切な回答のアルファベットを直接記入してください。}''
(\textit{Answer with the option's letter from the given choices directly.}
\\\\
\textbf{JMMMU-Pro:}\\
``\textit{あなたの役割は画像で与えられた問題を解くことです。与えられた選択肢の中から最も適切な回答のアルファベットを直接記入してください。}'' (\textit{Your role is to solve the question provided in the image. Answer with the option's letter from the given choices directly.})
\end{promptbox}

\begin{promptbox}[Inference Prompts: CoT]{darkblue}
\textbf{JMMMU:}\\
      ``\textit{次の選択式問題に答えてください。あなたの回答の最終行は、次の形式にしてください：'答え: \$LETTER'（引用符なし）。ここで LETTER は選択肢のいずれかの文字です。答える前に step by step で考えてください。}'' (\textit{Answer the following multiple choice question. The last line of your response should be of the following format: 'Answer: \$LETTER' (without quotes) where LETTER is one of options. Think step by step before answering.})
(\textit{}
\\\\
\textbf{JMMMU-Pro:}\\
``\textit{あなたの役割は画像で与えられた問題を解くことです。あなたの回答の最終行は、次の形式にしてください：'答え: \$LETTER'（引用符なし）。ここで LETTER は選択肢のいずれかの文字です。答える前に step by step で考えてください。}'' (\textit{Your role is to solve the question provided in the image. The last line of your response should be of the following format: 'Answer: \$LETTER' (without quotes) where LETTER is one of options. Think step by step before answering.})
\end{promptbox}

\begin{promptbox}[Inference Prompts: OCR Task]{orange}
\textbf{OCR Task Prompt:}\\
``\textit{画像内の選択式（多肢選択式）問題から、導入説明文を含む 質問文全体 と、対応する 選択肢 を抽出して出力してください。関連画像のテキストや問題番号は除外してください。OCR のみを行い、問題を解こうとしないでください。Formatは, 'Question: {}, Options: {}' (引用符なし)の形で出力してください}'' (\textit{Extract and output the full text of the question, including any introductory descriptions, as well as the corresponding answer choices from the multiple-choice question in the image. Exclude any text from associated images or the question number. Perform OCR only; do not attempt to solve the question. Please output in the format: 'Question: {}, Options: {}' (without quotes).})
\end{promptbox}

\section{Full Results}
\label{sec_appendix:full_results}
In \Cref{table:main_results_direct} and \Cref{table:main_results_cot}, we present detailed results for both the Direct Prompt and the CoT Prompt.

\begin{table*}[h]
\centering
\centering
\begin{tabular}{@{}lc|c|cc|cc@{}}
\toprule
Model & JMMMU-Pro & \color{gray}JMMMU & CS Pro & \color{gray}CS & CA Pro & \color{gray}CA \\
 & (1320) & \color{gray}(1320) & (600) & \color{gray}(600) & (720) & \color{gray}(720) \\
\midrule
\textbf{Random} & & & & & &  \\
~~~~Random Choice & 27.05 & \color{gray}27.05 & 26.33 &  \color{gray}26.33 & 27.64  & \color{gray}27.64 \\
~~~~Frequent Choice  & 27.73 & \color{gray}27.73 & 25.33 &  \color{gray}25.33 & 29.72 & \color{gray}29.72 \\
\midrule
\textbf{Multilingual Open LMMs} & & & & & &  \\
~~~~Qwen3-VL-8B & 45.83 & \color{gray}46.82 & 47.00 & \color{gray}56.33 & 44.86 & \color{gray}38.89 \\
~~~~Qwen2.5-VL-7B & 44.70 & \color{gray}46.82 & 50.17 & \color{gray}57.83 & 40.14 & \color{gray}37.64 \\
~~~~Phi-4-multimodal & 31.82 & \color{gray}39.55 & 28.83 & \color{gray}38.00 & 34.31 & \color{gray}40.83 \\
~~~~Aya-Vision-8B & 22.42 & \color{gray}32.05 & 23.83 & \color{gray}40.67 & 21.25 & \color{gray}24.86 \\
~~~~Pangea-7B & 19.55 & \color{gray}37.50 & 23.00 & \color{gray}47.17 & 16.67 & \color{gray}29.44 \\
\midrule
\textbf{English-centric Open LMMs} & & & & & &  \\
~~~~LLaVA-OV-1.5-8B & 29.92 & \color{gray}51.74 & 26.33 & \color{gray}53.33 & 32.92 & \color{gray}50.42 \\
~~~~LLaVA-OV-7B & 27.35 & \color{gray}41.14 & 26.50 & \color{gray}43.83 & 28.06 & \color{gray}38.89 \\
~~~~InternVL2.5-8B & 25.08 & \color{gray}41.36 & 23.83 & \color{gray}43.33 & 26.11 & \color{gray}39.72 \\
\midrule
\textbf{Japanese Open LMMs} & & & & & &  \\
~~~~Sarashina2.2-V-3B & 38.03 & \color{gray}47.95 & 40.17 & \color{gray}61.50 & 36.25 & \color{gray}36.67 \\
~~~~Sarashina2-V-14B & 30.68 & \color{gray}37.27 & 32.33 & \color{gray}43.17 & 29.31 & \color{gray}32.36 \\
~~~~Sarashina2-V-8B & 27.88 & \color{gray}39.62 & 27.00 & \color{gray}51.00 & 28.61 & \color{gray}30.14 \\
~~~~Heron-NVILA-Lite-15B & 26.97 & \color{gray}50.15 & 26.67 & \color{gray}59.17 & 27.22 & \color{gray}42.64 \\
\midrule
\textbf{Closed LMMs} & & & & & &  \\
~~~~Gemini3Pro (reasoning high) & 87.04 & \color{gray}89.77 & 95.00 & \color{gray}95.00 & 80.42 & \color{gray}85.42 \\
~~~~GPT-5.2 (reasoning high) & 83.33 & \color{gray}84.47 & 88.33 & \color{gray}85.50 & 79.17 & \color{gray}83.61 \\
\bottomrule
\end{tabular}
\caption{\textbf{Results with the direct prompt.}}
\label{table:main_results_direct}
\end{table*}

\begin{table*}[h]
\centering
\centering
\begin{tabular}{@{}lc|c|cc|cc@{}}
\toprule
Model & JMMMU-Pro & \color{gray}JMMMU & CS Pro & \color{gray}CS & CA Pro & \color{gray}CA \\
 & (1320) & \color{gray}(1320) & (600) & \color{gray}(600) & (720) & \color{gray}(720) \\
\midrule
\textbf{Random} & & & & & &  \\
~~~~Random Choice & 27.05 & \color{gray}27.05 & 26.33 &  \color{gray}26.33 & 27.64  & \color{gray}27.64 \\
~~~~Frequent Choice  & 27.73 & \color{gray}27.73 & 25.33 &  \color{gray}25.33 & 29.72 & \color{gray}29.72 \\
\midrule
\textbf{Multilingual Open LMMs} & & & & & &  \\
~~~~Qwen3-VL-8B & 47.27 & \color{gray}52.88 & 47.50 & \color{gray}55.83 & 47.08 & \color{gray}50.42 \\
~~~~Qwen2.5-VL-7B & 45.00 & \color{gray}47.65 & 46.67 & \color{gray}54.00 & 43.61 & \color{gray}42.36 \\
~~~~Phi-4-multimodal & 24.17 & \color{gray}32.05 & 22.00 & \color{gray}31.50 & 25.97 & \color{gray}32.50 \\
~~~~Aya-Vision-8B & 26.74 & \color{gray}37.73 & 27.00 & \color{gray}40.33 & 26.53 & \color{gray}35.56 \\
~~~~Pangea-7B & 23.41 & \color{gray}34.09 & 21.67 & \color{gray}36.17 & 24.86 & \color{gray}32.36 \\
\midrule
\textbf{English-centric Open LMMs} & & & & & &  \\
~~~~LLaVA-OV-1.5-8B & 31.97 & \color{gray}46.44 & 28.00 & \color{gray}46.83 & 35.28 & \color{gray}46.11 \\
~~~~LLaVA-OV-7B & 14.09 & \color{gray}21.29 & 14.33 & \color{gray}18.00 & 13.89 & \color{gray}24.03 \\
~~~~InternVL2.5-8B & 31.21 & \color{gray}34.32 & 29.00 & \color{gray}39.67 & 33.06 & \color{gray}29.86 \\
\midrule
\textbf{Japanese Open LMMs} & & & & & &  \\
~~~~Sarashina2.2-V-3B & 42.88 & \color{gray}45.30 & 54.00 & \color{gray}59.00 & 33.61 & \color{gray}33.89 \\
~~~~Sarashina2-V-14B & 30.00 & \color{gray}35.00 & 30.50 & \color{gray}44.50 & 29.58 & \color{gray}27.08 \\
~~~~Sarashina2-V-8B & 27.27 & \color{gray}34.39 & 25.33 & \color{gray}40.83 & 28.89 & \color{gray}29.03 \\
~~~~Heron-NVILA-Lite-15B & 5.30 & \color{gray}34.70 & 1.00 & \color{gray}38.33 & 8.89 & \color{gray}31.67 \\ \\
\bottomrule
\end{tabular}
\caption{\textbf{Results with the CoT prompting.}}
\label{table:main_results_cot}
\end{table*}

\section{Failure Cases}
\label{sec_appendix:failure_cases}
\begin{figure*}[t]
\centering
    \includegraphics[width=0.99\linewidth]{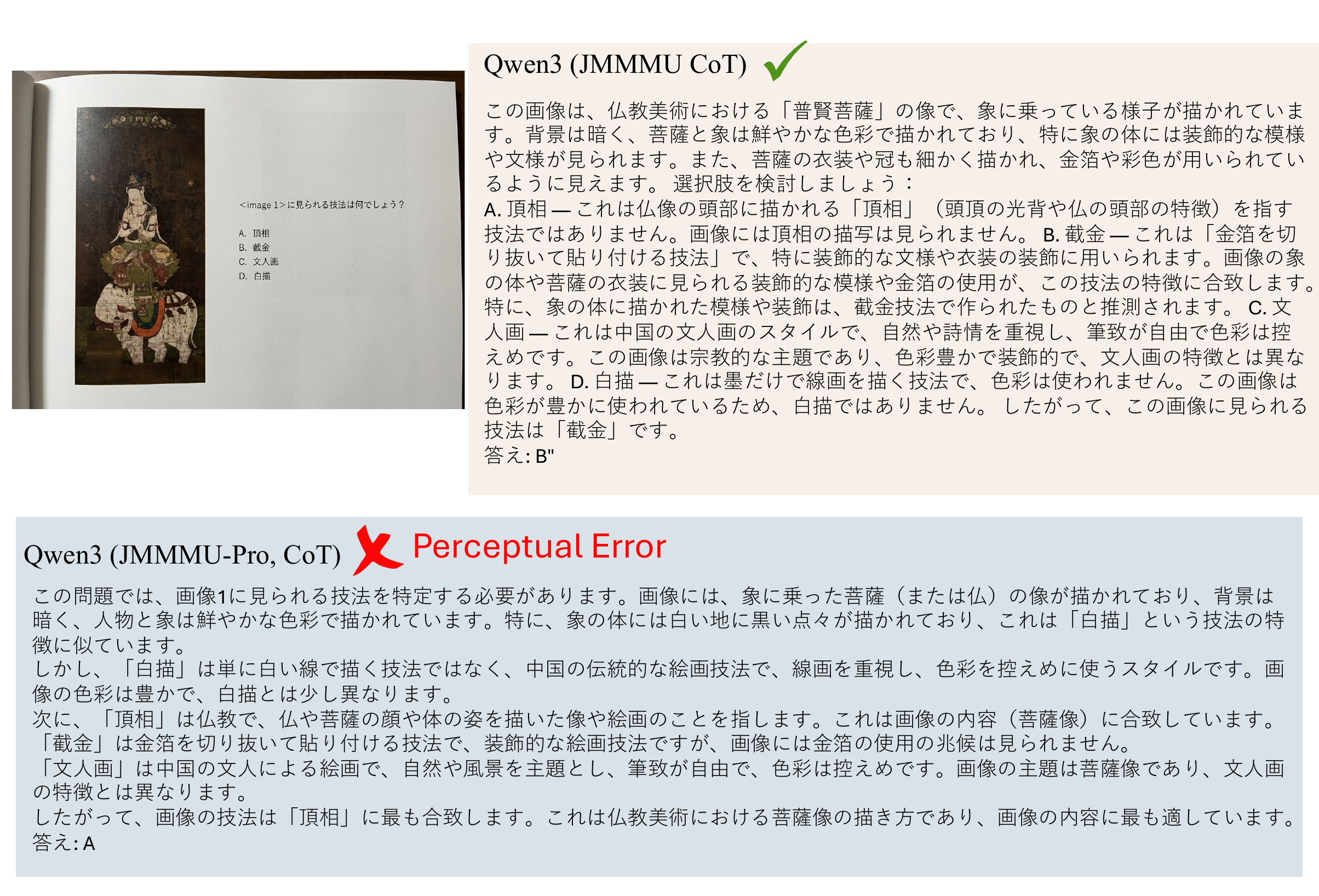}\\
   \caption{Failure case: Perceptual error.}
    \label{fig:failure_1}
\end{figure*}

\begin{figure*}[t!]
\centering
    \includegraphics[width=0.80\linewidth]{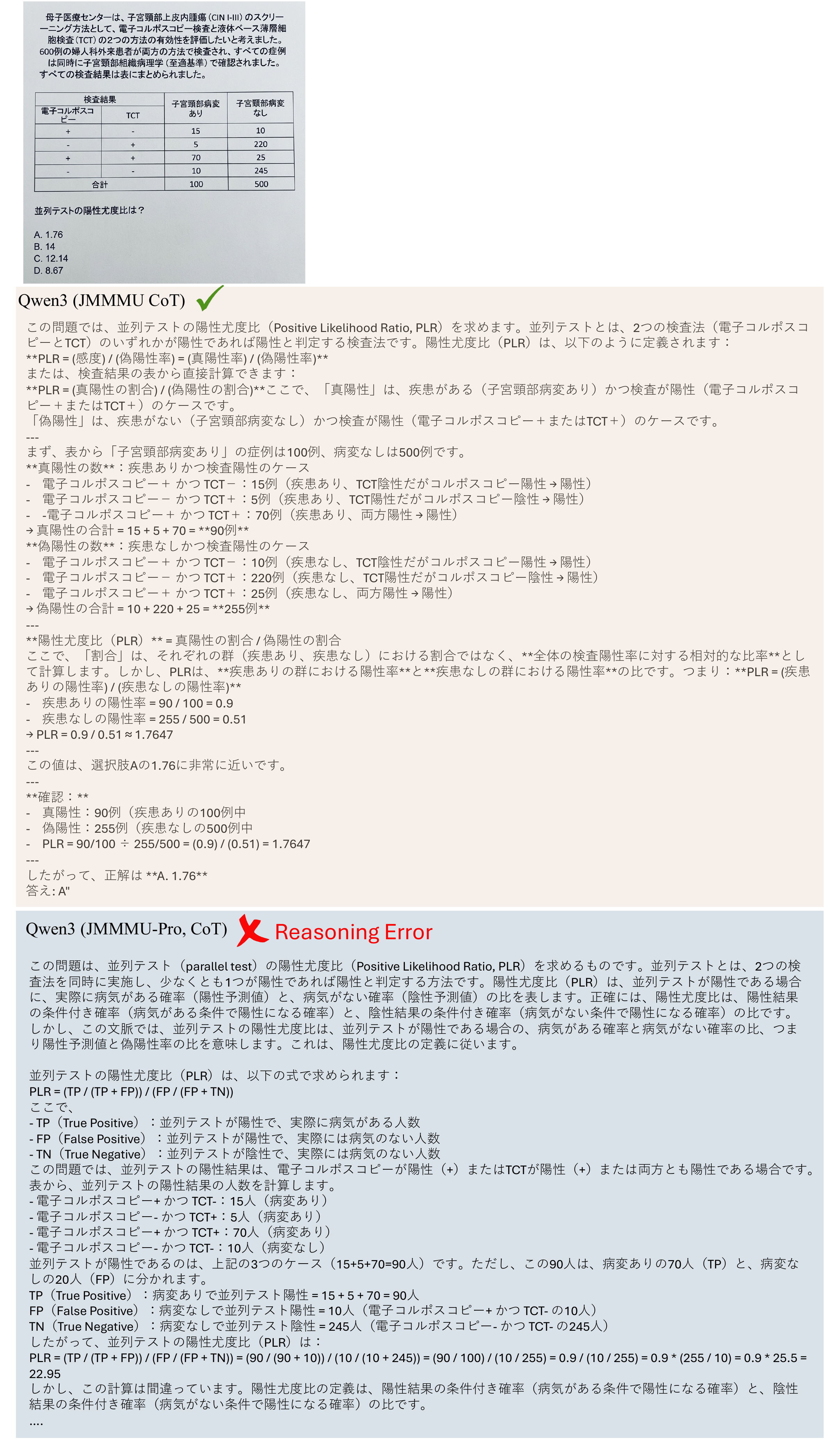}\\
   \caption{Failure case: Reasoning error.}
    \label{fig:failure_2}
\end{figure*}

We show representative failure cases of the state-of-the-art open-source LMM, Qwen3-VL-8B, in \Cref{fig:failure_1} and \Cref{fig:failure_2}. These examples show that the model exhibits both reasoning errors and perceptual errors only in JMMMU-Pro, indicating that JMMMU-Pro requires deeper integrated visual–textual understanding beyond mere OCR capability.

\end{document}